\documentclass[letterpaper, 10pt, conference]{ieeeconf}

\IEEEoverridecommandlockouts                              

\usepackage{pgfplots}
\pgfplotsset{compat=newest}
\usetikzlibrary{plotmarks}
\usetikzlibrary{arrows.meta}
\usepgfplotslibrary{patchplots}
\usepackage[noadjust]{cite}
\usepackage{grffile}
\usepackage{multicol}
\usepackage[fleqn]{amsmath}
\usepackage{amssymb}
\usepackage{comment}
\usepackage{color}
\usepackage{graphicx}
\usepackage{pstricks}
\usepackage{mathtools}
\usepackage[ruled,linesnumbered]{algorithm2e}
\usepackage{setspace}
\usepackage{breqn}
\usepackage{makecell}
\usepackage{multirow}
\usepackage{afterpage}
\usepackage{subcaption}
\usepackage{siunitx}
\usepackage{import}
\makeatletter
\let\NAT@parse\undefined
\makeatother
\usepackage[hidelinks,bookmarks=true]{hyperref}

\allowdisplaybreaks

\newcommand{\argmin}{\operatornamewithlimits{argmin}}
\newcommand{\argmax}{\operatornamewithlimits{argmax}}
\newcommand{\norm}[1]{\left\lVert #1 \right\rVert}
\def\argmax{\mathop{\rm argmax}}
\def\argmin{\mathop{\rm argmin}}

\def\atan2{\mathop{\rm atan2}}

\title{\LARGE \bf
A Maximum Likelihood Approach to Extract Polylines\\ 
from 2\nobreakdash-D Laser Range Scans}

\author{Alexander~Schaefer, Daniel~B\"uscher, Lukas~Luft, Wolfram~Burgard
\thanks{\copyright\ 2018 IEEE. Personal use of this material is permitted.  Permission from IEEE must be obtained for all other uses, in any current or future media, including reprinting/republishing this material for advertising or promotional purposes, creating new collective works, for resale or redistribution to servers or lists, or reuse of any copyrighted component of this work in other works.}
\thanks{This work has been partially supported by the European Commission in the Horizon 2020 framework program under grant agreement 645403-RobDREAM, and by Samsung Electronics~Co.~Ltd. under the GRO program.}%
\thanks{All authors are with the Department of Computer Science, University of Freiburg, Germany.}%
\thanks{\tt \small \{aschaef, buescher, luft, burgard\} @cs.uni-freiburg.de}}%

\begin{document}

\maketitle
\thispagestyle{empty}
\pagestyle{empty}

\begin{abstract}
Man-made environments such as households, offices, or factory floors are typically composed of linear structures. 
Accordingly, polylines are a natural way to accurately represent their geometry. 
In this paper, we propose a novel probabilistic method to extract polylines   from raw 2\nobreakdash-D laser range scans. 
The key idea of our approach is to determine a set of polylines that maximizes the likelihood of a given scan.  
In extensive experiments carried out on publicly available real-world datasets and on simulated laser scans, we demonstrate that our method substantially outperforms existing state-of-the-art approaches in terms of accuracy, while showing comparable computational requirements.
Our implementation is available under \url{https://github.com/acschaefer/ple}.
\end{abstract}

\section{Introduction}
\label{sec:introduction}

In order to navigate planar, structured environments like offices, households, or factory work floors, mobile robots often rely on horizontally mounted 2\nobreakdash-D laser scanners.
These sensors allow them to create floor plan-like maps, which they in turn use to localize themselves.
Mapping and localization based on raw laser data, however, demand large amounts of computation power and memory, both of which tend to be restricted on a mobile platform.  

\begin{figure}
	\centering
	\resizebox{\columnwidth}{!}{\huge 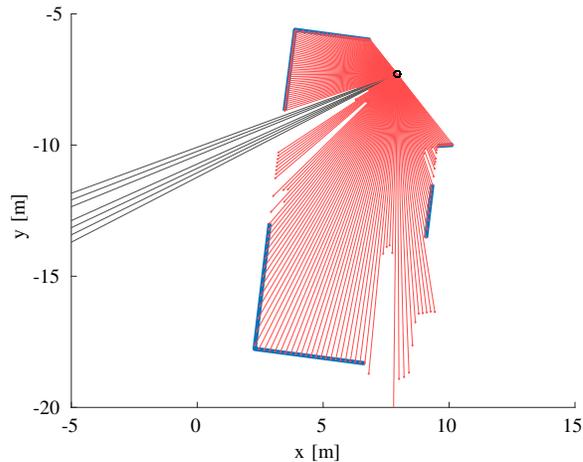}
	\caption{Exemplary result of our polyline extraction method applied to a scan captured in an office.  
		The original scan consists of \num{361} rays, of which every second is displayed as a red line. 
		Gray lines indicate maximum-range readings.  
		The extracted polyline map, drawn as blue lines, consists of only ten vertices, reducing memory requirements to less than \SI{3}{\%}.  
		On average, the distance between the measured endpoints of the rays and their hypothetical intersections with the map is as low as~\SI{5.4}{mm}, with a maximum absolute value of~\SI{23.0}{mm}.}
	\label{fig:intro_scan}
\end{figure}

A popular solution to this problem is feature extraction.  
Encoding all the data of a scan in a few polyline features, for example, can drastically reduce computation time and memory footprint.
This is due to the ability of polylines to exploit the high redundancy of scans recorded in approximately line-shaped environments.
Consider figure~\ref{fig:intro_scan}, which depicts a typical laser scan captured in an office.  
The scan spends hundreds of rays to describe straight walls, while a set of polylines with a total of ten vertices is sufficient to accurately represent these linear structures.

Polyline features like the ones in figure~\ref{fig:intro_scan} have been shown to be useful for different tasks in mapping and localization, for example for feature-based SLAM~\cite{castellanos1999, garulli2005, rodriguez2006, choi2008, lv2014} or for tracking line segments in consecutive 2\nobreakdash-D scans recorded by a moving sensor to estimate 3\nobreakdash-D planes in the environment~\cite{berger2013}.

In this paper, we present a novel method to extract polylines from 2\nobreakdash-D laser scans.
What sets our approach apart from most others is its probabilistic motivation.
The derived algorithm does not rely on a geometric heuristic, but strives to find the set of polylines that maximizes the measurement likelihood of the scan.
Furthermore, while most other approaches operate on the scan endpoints only and thereby discard valuable information encoded in the ray paths, our algorithm leverages this data to yield as accurate polyline estimates as possible.
As demonstrated in our experimental evaluation, this results in superior accuracy both on real-world and on simulated data.

\section{Related Work}
\label{sec:related_work}

In this section, we provide an overview over existing work on feature extraction techniques for 2\nobreakdash-D lidar scans and related sensor modalities.

Early approaches to extract line features from 2\nobreakdash-D laser scans produce so-called line maps~\cite{sack2004}.
As opposed to the chains of line segments that are polylines, line maps model the environment by infinite lines, and thus suffer from two major drawbacks:
They can only represent convex maps, and they do not store any topology information, i.e. the connections between the lines.
This is why infinite line representations have largely been replaced by polylines.
For a survey on different methods to build infinite line maps, see Sack and Burgard~\cite{sack2004}.

Nguyen et~al.~\cite{nguyen2005} present an overview of various techniques to extract line segments from 2\nobreakdash-D lidar data and evaluate the performance of six popular algorithms experimentally.
They conclude that split-and-merge~\cite{pavlidis1974} and iterative endpoint fit~\cite{douglas1973} perform most favorably in terms of accuracy and speed.

Each of the following approaches tackles the line extraction problem from planar scans from a different perspective.
Borges and Aldon~\cite{borges2000} use a fuzzy clustering algorithm, which does not require prior knowledge of the number of lines.
Latecki and Lakaemper~\cite{latecki2006} combine perceptual grouping techniques with the expectation maximization algorithm to determine polylines.
Pfister and Burdick~\cite{pfister2006} extract both line and point features from scan endpoints using a multi-scale Hough transform.
Similarly, Berrio et~al.~\cite{berrio2012} determine line segments via the Hough transform in combination with a so-called successive edge following algorithm.
Harati and Siegwart~\cite{harati2007} build a wavelet framework to extract initial estimates of line segments.
They do not, however, provide the corresponding line fitting algorithm.

In cartography, the Visvalingam line simplification algorithm~\cite{visvalingam1993} is a popular method to reduce the numbers of vertices of a polyline by iteratively removing the vertex that incurs the least perceptible change.
Although to our knowledge, the algorithm has not reportedly put to use for line extraction from laser scans, it is well suited for this task.

All methods discussed thus far have in common that they are built upon some kind of heuristic and lack a probabilistic foundation.
In contrast to those, Pfister et~al.~\cite{pfister2003} present a take on line extraction from multiple scans that follows an elaborate maximum likelihood formalism.
First, they generate a set of infinite line estimates using the Hough transform.
Then, taking into account the pose uncertainty of the robot and the measurement uncertainty of the sensor, they numerically maximize the measurement likelihood over the line parameters.
In order to obtain line segments, they finally project the scan points onto the infinite lines and crop them accordingly.
As opposed to this method, our method leverages probabilistics not only to optimize a given initial line estimate, but also to generate the estimate itself.

In another probabilistic approach, Veeck and Burgard~\cite{veeck2004} formulate an algorithm to extract polylines from multiple 2\nobreakdash-D laser scans captured at known poses.
In the first step, they create an occupancy grid map, which they then use to estimate the line contours of the environment.
Second, they repeatedly apply a set of eight operations to these initial lines, including merging, splitting, adding vertices, removing vertices, moving vertex locations on a raster grid, and removing the resulting zig-zag patterns.
In this way, they strive to optimize the Euclidean distances between the laser scan endpoints and the nearest polyline.
Our approach is different from Veeck and Burgard's in various aspects.
For example, it is less complicated both conceptually and in terms of implementation.
Moreover, their approach does not incorporate any ray path information in the result.

Polylines are useful features not only in the context of lidar.
For an approach to extract line segments from sonar data using the Hough transform, see Tard\'os et~al.~\cite{tardos2002}.
Navarro et~al. present methods to localize a robot using lines extracted from a rotating ultrasound sensor~\cite{navarro2007} and from infrared distance sensors~\cite{navarro2008}.

For 2\nobreakdash-D laser scans, there is only little research investigating features other than lines.
Tipaldi and Arras' multi-scale FLIRT descriptors~\cite{tipaldi2010} are among these few.
Bosse and Zlot convert a whole laser scan into a single feature \cite{bosse2008} and extract features from quadratic areas formed by a set of scans \cite{bosse2009}.

\section{Approach}
\label{sec:approach}

In this work, we present a method to extract a set of polyline features from a 2\nobreakdash-D lidar scan.
In contrast to prevalent line extraction techniques like split-and-merge or iterative endpoint fit, our approach does not rely on a geometric heuristic, but maximizes the measurement probability of the scan to accurately determine polylines.

The method consists of two steps: polyline extraction and polyline optimization.
Polyline extraction starts by connecting all neighboring scan endpoints to form a set of initial polylines.
It then iteratively removes the vertex that incurs the least error in terms of measurement probability until it reaches a given threshold.
The result is a set of polylines whose vertex locations coincide with the locations of a subset of the scan endpoints.
To do away with this limitation, we formulate an optimization problem that moves the vertices to the positions that maximize the measurement probability of the scan.
We call this latter process polyline optimization.

In the following, we first define the probabilistic sensor model.
Then, we explain the polyline extraction step, before going into the details of polyline optimization.

\subsection{Probabilistic Sensor Model}

Both polyline extraction and polyline optimization strive to maximize the measurement probability.
By measurement probability, we refer to the probability of a laser scan conditioned on a specific set of polylines.
In order to describe this quantity mathematically, we need to define all necessary variables.
We denote the scan by~\mbox{$Z \coloneqq \{z_k\}$}, where \mbox{$k \in \{1,2,\ldots,K\}$} represents the index of the laser ray.
A single laser measurement~\mbox{$z \coloneqq \{a,b\}$} is composed of two two-element column vectors: the starting point of the ray~$a$ and the endpoint~$b$.
The set of polylines~$L$ consists of a total of $I$ individual polylines.
These polylines are ordered sets  $l \coloneqq \{v_j\}$, composed of at least two pairwise distinct vertices~$v$.
The vertices~$v$, just like the ray starting points and endpoints $a$ and $b$, are specified with respect to the coordinate system of the polyline map~$L$.
Note that no vertex can be part of multiple polylines:
\begin{equation*}
\bigcap\limits_{i=1}^I l_i = \emptyset.
\end{equation*}

Most lidar sensors exhibit approximately normally distributed noise in radial direction and relatively small angular noise.
Consequently, we neglect angular noise and model the distribution of the measured ray radius given the polyline map as a Gaussian probability density function centered at the true radius.
With the above definitions, we formulate the sensor model for a single ray as
\begin{equation}
\label{eq:sensor_model_ray}
p(z \mid L)
	= \mathcal{N}(r(z); \hat{r}(z,L), \Sigma),
\end{equation}
with the measured ray radius \mbox{$r(z) \coloneqq \norm{b-a}$}.
The function $\hat{r}(z,L) \in \mathbb{R}^+$ computes the distance between the starting point of the ray and the first intersection between its axis and the polyline map.
The variance of the radial noise~$\Sigma$ is usually a function of multiple parameters such as the sensor device, the ray radius, the optical properties of the reflecting surface, and temperature.
It can either be read off the datasheet of the sensor or determined experimentally.

By assuming independence between the individual laser rays of a scan, we extend equation~\eqref{eq:sensor_model_ray} to compute the measurement probability of the whole scan as
\begin{equation*}
\label{eq:sensor_model_scan}
p(Z \mid L) 
	= \prod_{k=1}^K p(z_k \mid L).
\end{equation*}
This formula represents the measurement probability that both steps of our algorithm strive to maximize.

\subsection{Polyline Extraction}
\label{sec:polyline_extraction}

Line extraction is always a compromise between memory requirements and accuracy of the produced lines.
This compromise needs to be quantified.
Embedded applications, for example, might focus on minimal memory footprint, while offline mapping systems might favor high accuracy at the expense of polylines that consist of hundreds or thousands of vertices.
For this reason, every line extraction algorithm requires some kind of parameter.
In the following, we choose this parameter to be the maximum number of vertices $J_{\max}$ of the polyline set, because it allows direct control over the memory footprint of the result.
Note, however, that our approach makes it easy to use arbitrary parameters, for example the maximum root mean squared error of the ray radii, the Akaike Information Criterion, the maximum difference in area between the extracted polylines and the polygon of the original scan endpoints, etc., as described further below.

Given a specific maximum number of vertices~$J_{\max}$, the goal of the polyline extraction step is to find the set of polylines~$L^*$ with at most $J_{\max}$ vertices that, among all other polyline maps with at most $J_{\max}$ vertices, yields the highest measurement probability.
Formally, we are confronted with the optimization problem
\begin{equation}
\label{eq:optimization_problem}
L^* = \argmax_L p(Z \mid L)~\Big|~J(L) = J_{\max},
\end{equation}
where $J(L)$ denotes the number of vertices in $L$.

Solving~\eqref{eq:optimization_problem} is primarily a combinatorial problem.
Even if we knew the locations of the $J_{\max}$ vertices, we still do not know the data associations, i.e. which vertices make up which polyline.
Exhaustively searching the space of all data associations for the combination that maximizes the measurement probability quickly leads to combinatorial explosion even for small~$J_{\max}$.
For that reason, we use a greedy algorithm to solve the combinatorial part of $\eqref{eq:optimization_problem}$.

In a nutshell, the algorithm first creates a polygon by connecting all neighboring scan endpoints.
Starting from this initial map, it iteratively removes the vertex that reduces the measurement probability of the scan given the map by the least amount, until it reaches the desired number of vertices, or until another stopping criterion like one of those mentioned above is fulfilled.

\begin{figure}
	\centering
	\includegraphics{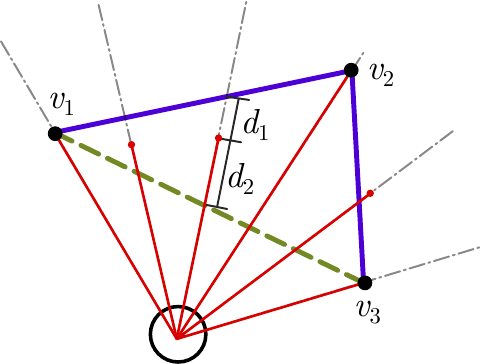}
	\caption{Illustration of the different terms in equation~\eqref{eq:min_error}.
		The black circle represents the lidar sensor.
		It shoots laser rays in different directions, drawn as red lines with dots marking their measured endpoints.
		The blue polyline is the map~$L$, its vertices are denoted $v_1$, $v_2$, $v_3$.
		The green dashed line depicts~$L\setminus v_2$, the polyline~$L$ with its middle vertex~$v_2$ removed.
		When computing the error~$e_2$ corresponding to the removal of $v_2$, we need to determine two quantities for each ray~$z_k$.
		The first one is the distance~$d(z_k,L)$ between the measured endpoint and the intersection between the ray and the given polyline~$L$, exemplarily shown for one ray and denoted $d_1$ in the graphic.
		The second one is the distance~$d(z_k,L\setminus v_2)$ between the endpoint and the polyline with the vertex~$v_2$ removed, denoted $d_2$ in the graphic.
	}
	\label{fig:vertex_error}
\end{figure}

Given the initial or any intermediate polyline map~$L$, the problem of finding the vertex~$v_{j^*}$ that reduces the measurement probability the least can be formulated as
\begin{align}
\label{eq:min_error}
j^* 
	&= \argmax_j p(Z \mid L \setminus v_j) \notag \\
	&= \argmax_j \log\{p(Z \mid L \setminus v_j)\} \notag \\
	&= \argmin_j \sum_{k=1}^K 
		\frac{d^2(z_k,L\setminus v_j)}{\Sigma_k} \notag \\
	&= \argmin_j \Bigg\{ \sum_{k=1}^K \frac{d^2(z_k,L\setminus v_j)}{\Sigma_k} 
		- \sum_{k=1}^K \frac{d^2(z_k,L)}{\Sigma_k} \Bigg\} \notag \\
	&= \argmin_j \sum_{k=1}^K 
		\frac{d^2(z_k,L\setminus v_j) - d^2(z_k,L)}{\Sigma_k} \notag \\
	&= \argmin_j \underbrace{ \sum_{k \in X(L,v_j)}
		\frac{d^2(z_k,L\setminus v_j) - d^2(z_k,L)}{\Sigma_k}}_{\eqqcolon e_j} \notag \\
	&= \argmin_j \{e_j\}, 
\end{align}
where we define \mbox{$L \setminus v_j \coloneqq \{l_i \setminus v_j\} \mid i = 1,2,\ldots,I$}.
Accordingly, \mbox{$p(Z \mid L \setminus v_j)$} represents the probability density function of the measurements conditioned on the set of polylines with the vertex~$v_j$ removed.
The function \mbox{$d(z,L) \in \mathbb{R}$} determines the distance between the endpoint of a ray and its intersection with the map
\begin{align*}
d(z,L) \coloneqq r(z) - \hat{r}(z,L).
\end{align*}
Please note that the transition from the third to the fourth line of equation~\eqref{eq:min_error} is valid because the second sum~\mbox{$\sum_{k=1}^K d^2(z_k,L)\,\Sigma_k^{-1}$} is constant with respect to $j$.
The function $X(L,v_j) \subseteq \{1,2,\ldots,K\}$ in the sixth line returns the indices of the rays that intersect any of the line segments that start or end at $v_j$.
The variable~$e_j$ can be thought of as a measurement probability error term corresponding to the removal of the $j$\nobreakdash-th vertex.
Consequently, removing the vertex that decreases the measurement probability the least is equivalent to removing the vertex whose removal incurs the smallest error.
For an illustration of the quantities that need to be determined in order to compute the errors, see figure~\ref{fig:vertex_error}.

Computing the error terms according to equation~\eqref{eq:min_error} is valid if the vertex in question has two neighbors.
If the vertex marks the start or the end of the polyline, however, removing it means removing the corresponding line segment.
In this case, it is not possible to compute the term~$d(z_k,L\setminus v_j)$.
Consequently, we resort to a heuristic:
We introduce a constant parameter~$d_{\textrm{rm}}$, which acts as a placeholder for $d(z_k,L\setminus v_j)$ in case the latter term is impossible to determine.
The magnitude of $d_{\textrm{rm}}$ controls how easily the algorithm crops lines.
If chosen large, the algorithm rather keeps the line segments and prefers to remove the vertices in the middle of the polylines.
If chosen small, it tends to crop lines and reluctantly removes two-neighbor vertices.

\begin{algorithm}
	\setstretch{1.35}
	\KwData{$Z$, $r_{\max}$, $l_{\max}$, $J_{\max}$}
	\KwResult{$L$}
	$L \gets \{\} $ \\
	\For{all $z_k$ in $Z$}
	{
		\If{$\big(\max(r_k,r_{k+1}) \leq r_{\max}\big)$ \\
			$\land~\big(\norm{b_{k+1}-b_k} \leq l_{\max}\big)$}
		{
			$l \gets \{b_k,b_{k+1}\}$ \\
			add $l$ to $L$ \\
		}
	}
	merge all line segments $l \in L$ to polylines so that \\
		\qquad $\bigcap\limits_{i=1}^I l_i = \emptyset$ \\
	$E \gets \{\}$\\
	\For{all $v_j$ in $L$}
	{
		compute error $e_j$ corresponding to removal of $v_j$ \\
		add $e_j$ to $E$ \\
	}
	\While{$J(L) > J_{\max}$}
	{
		find index $j^*$ of smallest element in $E$ \\
		remove $v_{j^*}$ from $L$ \\
		remove $e_{j^*}$ from $E$ \\
		recompute errors $e_{j^*-1}$ and $e_{j^*}$ in $E$ \\
	}
	\caption{Polyline Extraction}
	\label{algo:polyline_extraction}
\end{algorithm}

The pseudocode in listing~\ref{algo:polyline_extraction} shows the workings of the algorithm in detail and delineates an efficient implementation.
In lines~1 to 10, the algorithm forms the initial map.
To that end, it connects all neighboring scan endpoints that satisfy two conditions.
First, neither of the points in the pair that is to be connected is a maximum range reading (line~3).
Maximum-range readings emerge when there is no object within the range of the lidar sensor, so removing the corresponding line segments is consequential.
Second, the length of the connection between the points does not exceed a given maximum~$l_{\max}$ (line~4).
This step prevents the generation of long lines that are not sufficiently backed up by lidar data, for example connections between a short-range endpoint and a long-range endpoint, or connections between neighboring endpoints at large radii, far away from the sensor.
After the resulting set of line segments have been merged (lines~9 to 10), $L$ is either a polygon, if all endpoint pairs meet both conditions, or a set of polylines otherwise.

To reduce the number of vertices, the map~$L$ is then subjected to the greedy part of the algorithm, represented by lines~11 to 21.
Lines~11 to 15 initially compute the error values corresponding to the removal of the individual vertices $v_j$ in $L$.
The loop spanning lines~16 to 21 then iteratively removes the vertex corresponding to the smallest change in measurement probability and updates the errors, until the desired number of vertices is reached.
Note that in line~20, it is not necessary to recompute all errors in $E$.
Only the errors corresponding to the immediate neighbors of $v_{j^*}$ change.
After the removal of vertex $v_{j^*}$, those are indexed by $j^*-1$ and $j^*$.

If a stopping criterion other than the number of vertices is given, for example a maximum RMSE value, the condition in line~16 simply needs to be changed accordingly.

Algorithm~\ref{algo:polyline_extraction} solves the combinatorial part of the optimization problem~\eqref{eq:optimization_problem}.
It tells both which scan endpoints create which vertices, and which vertices form which polyline.
It does not, however, alter the positions of the vertices in order to maximize the measurement probability.
In polyline extraction, the vertex locations are limited to the locations of the scan endpoints.
The next step, polyline optimization, relaxes this restriction.

\subsection{Polyline Optimization}
\label{sec:polyline_optimization}

Having solved the combinatorial part of the optimization problem~\eqref{eq:optimization_problem}, we now turn to its numerical part:
We take the vertex locations produced by the polyline extraction step and move them to the positions that maximize the measurement probability of the scan conditioned on the map~\mbox{$p(Z \mid L)$}.

More formally, we want to solve
\begin{equation}
\label{eq:polyline_optimization}
L^* 
	= \argmax_L p(Z \mid L) 
	= \argmin_L \sum_{k=1}^K 
		\frac{d^2(z_k,L)}{\Sigma_k},
\end{equation}
which is a nonlinear, discontinuous, multivariate optimization problem in the coordinates of the polyline vertices.
Its discontinuous nature, which results from the polylines' kinks in the vertices, requires appropriate direct search solvers, for example the Nelder-Mead Simplex Method~\cite{lagarias1998}.

Before starting to optimize, it is important to closely consider the search space of the problem formulated in equation~\eqref{eq:polyline_optimization}.
In the case of a closed polygon, this space simply becomes $\mathbb{R}^{2J}$, where $J$ is the number of vertices in $L$.
A corresponding candidate solution consists of the coordinates of all polygon vertices.
In the case of a set of polylines, however, allowing all vertices to freely move around might lead to undesired effects:
Vertices at the start or end of a polyline might drift off into unobserved regions.
Consider vertices~$v_1$ and $v_3$ in figure~\ref{fig:vertex_error}, for example.
As long as they stay on the axis of the respective line segment, they can move indefinitely away from the observed region without affecting the measurement probability of the scan.
To avoid this effect and to keep the search space as small as possible, we constrain the movement of vertices at the start or end of a polyline to the axis of the corresponding ray.
Given a map consisting of $I$ polylines, the dimensionality of the search space hence becomes $2(J-I)$.

\section{Experiments}
\label{sec:experiments}

In order to assess the quality of the polyline maps produced by the presented method and to compare the results with those returned by existing approaches, we conduct two series of experiments.
In the first, we evaluate the performance of every method on real-world 2\nobreakdash-D lidar data.
The data is composed of \num{13} datasets taken from Radish, the publicly available Robotics Data Set repository~\cite{howard2003}.
From each of those datasets, listed in table~\ref{tab:datasets}, we randomly choose \num{20} scans, leading to a total of \num{260} scans to evaluate.
On average, each of the selected scans contains \num{264}~rays.
The second experiment series is based on the same number of simulated scans with \num{360}~rays each.
Simulation allows us to measure the accuracy of the obtained polyline maps not only with respect to the scan data, but also with respect to the underlying ground-truth map.
To simulate a scan, we first create a random polygon with \num{3}, \num{4}, \num{5}, \num{6}, \num{12}, \num{36}, or \num{180} vertices.
We then sample a noisy full-revolution, \num{360}-ray scan from it by first applying normally distributed noise to the ray angles, then computing the true intersection points of scan and polygon, and by finally adding normally distributed noise to the corresponding ray radii.
The standard deviations of angular and radial noise are \ang{0.2} and \SI{0.03}{m}, respectively.

\begin{table}[tp]
	\def\arraystretch{1.2}
	\caption{Datasets taken from the Robotics Data Set Repository~\cite{howard2003}.}
	\label{tab:datasets}
	\centering
	\begin{small}
	\begin{tabular}{|l|l|}
		\hline
		Acapulco Convention Center	& U Freiburg, 101 \\
		U Texas, ACES3  			& U Freiburg, campus \\ 
		Belgioioso Castle   		& MIT infinite corridor \\ 
		MIT, CSAIL          		& Intel Research Lab \\
		Edmonton Convention Center 	& \"Orebro University \\
		FHW museum          		& U Washington, Seattle \\
		U Freiburg, 079	&\\
		\hline
	\end{tabular}
	\end{small}
\end{table}

In both experiment series, we compare six different takes on polyline extraction, starting with Visvalingam line simplification~(VVL)~\cite{visvalingam1993}.
The method requires initial polylines, so we first connect the endpoints of the scan using the exact same procedure as described in algorithm~\ref{algo:polyline_extraction}, line~1 to 10, with $l_{\max}$ set to \SI{1}{m}.
Visvalingam's algorithm then simplifies the resulting initial polygon or set of polylines by iteratively removing the vertex whose removal is linked to the least perceptible change in the polyline.
The popular iterative endpoint fit algorithm~(IEF)~\cite{douglas1973} comes second in our comparison.
As opposed to the top-down approach of VVL, which starts with the most detailed line and iteratively simplifies it, IEF builds polylines from bottom up.
In short, IEF takes a set of range measurements, connects the first and the last point by a straight line, and then inserts the scan endpoint with the largest distance from the line as a vertex into the line.
It repeats this process until it reaches the specified number of vertices.
Split-and-merge~(SAM)~\cite{pavlidis1974} is an extension of IEF.
The only difference between the two algorithms is that in each iteration, SAM first fits the line estimate to the scan points by minimizing the squared distances between the points and the line.
Both algorithms do not account for maximum-range readings, which is why we removed them from the laser scans before passing the scans to IEF or SAM.
VB, the fourth method in our comparison, denotes the polyline learning algorithm proposed by Veeck and Burgard~\cite{veeck2004}.
We call our approach probabilistic line extraction (PLE).
If the vertices provided by PLE are optimized using the procedure described in section~\ref{sec:polyline_optimization}, we denote it by PLE+.
Throughout all experimental runs, we set \mbox{$l_{\max}=\SI{1}{m}$} and \mbox{$d_{\textrm{rm}}=\SI{0.5}{m}$}.
Furthermore, we assume that the radial variance~$\Sigma$ of all rays is the same.
As a consequence, we do not have to specify any variance at all, because if constant, the term~$\Sigma_k$ vanishes from equations~\eqref{eq:min_error} and \eqref{eq:polyline_optimization}.

Figure~\ref{fig:scan} exemplarily illustrates the results obtained by applying the described methods to the same scan.
Although the desired number of vertices was set to $J=10$ for all methods, the returned polyline maps differ considerably.

\def \scanfigwidth {0.24\linewidth}
\begin{figure*}
	\begin{subfigure}{\scanfigwidth}
		\resizebox{\columnwidth}{!}{\Huge 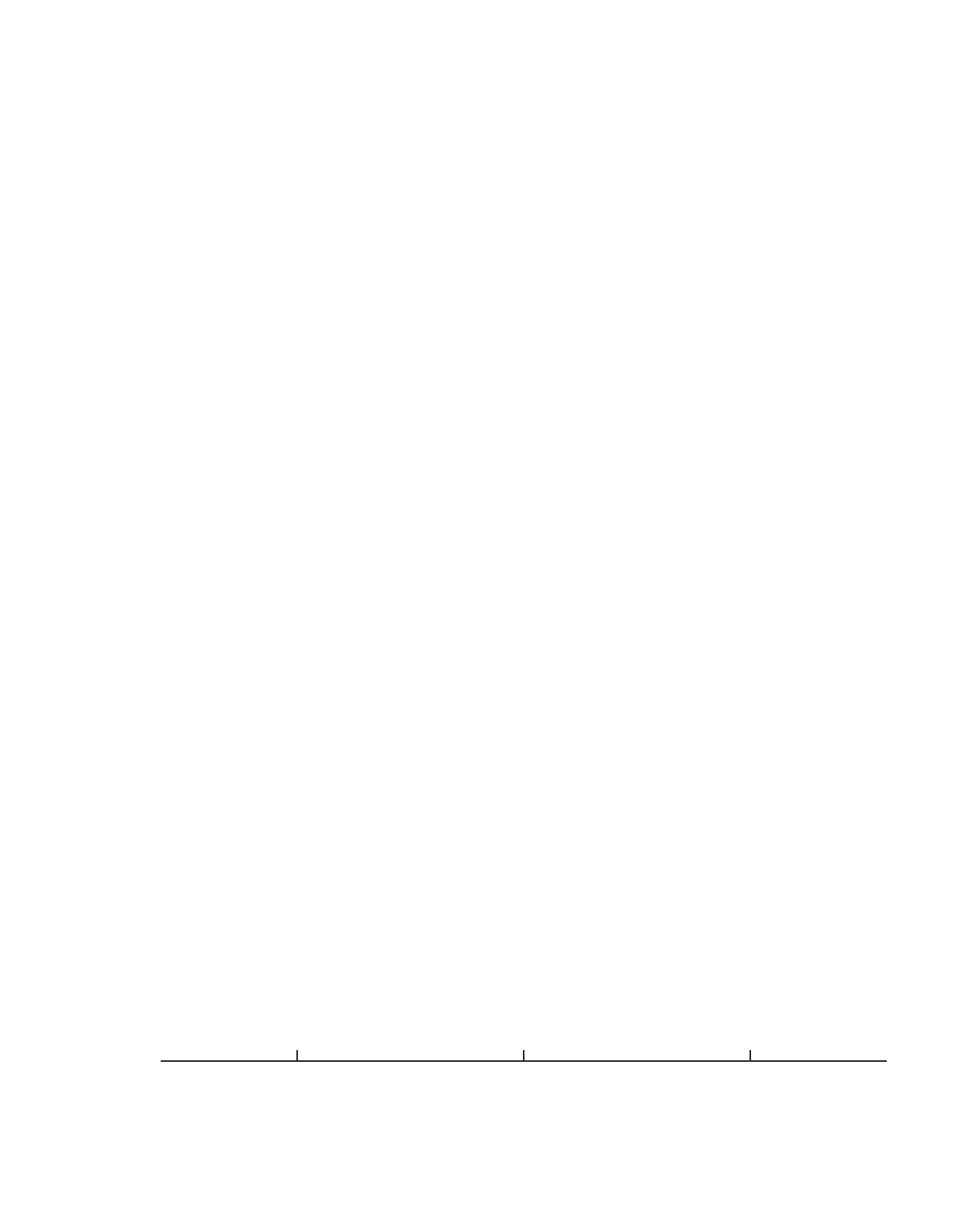}
		\caption{Visvalingam's algorithm.}
		\label{fig:vvl}
	\end{subfigure}
	\begin{subfigure}{\scanfigwidth}
		\resizebox{\columnwidth}{!}{\Huge 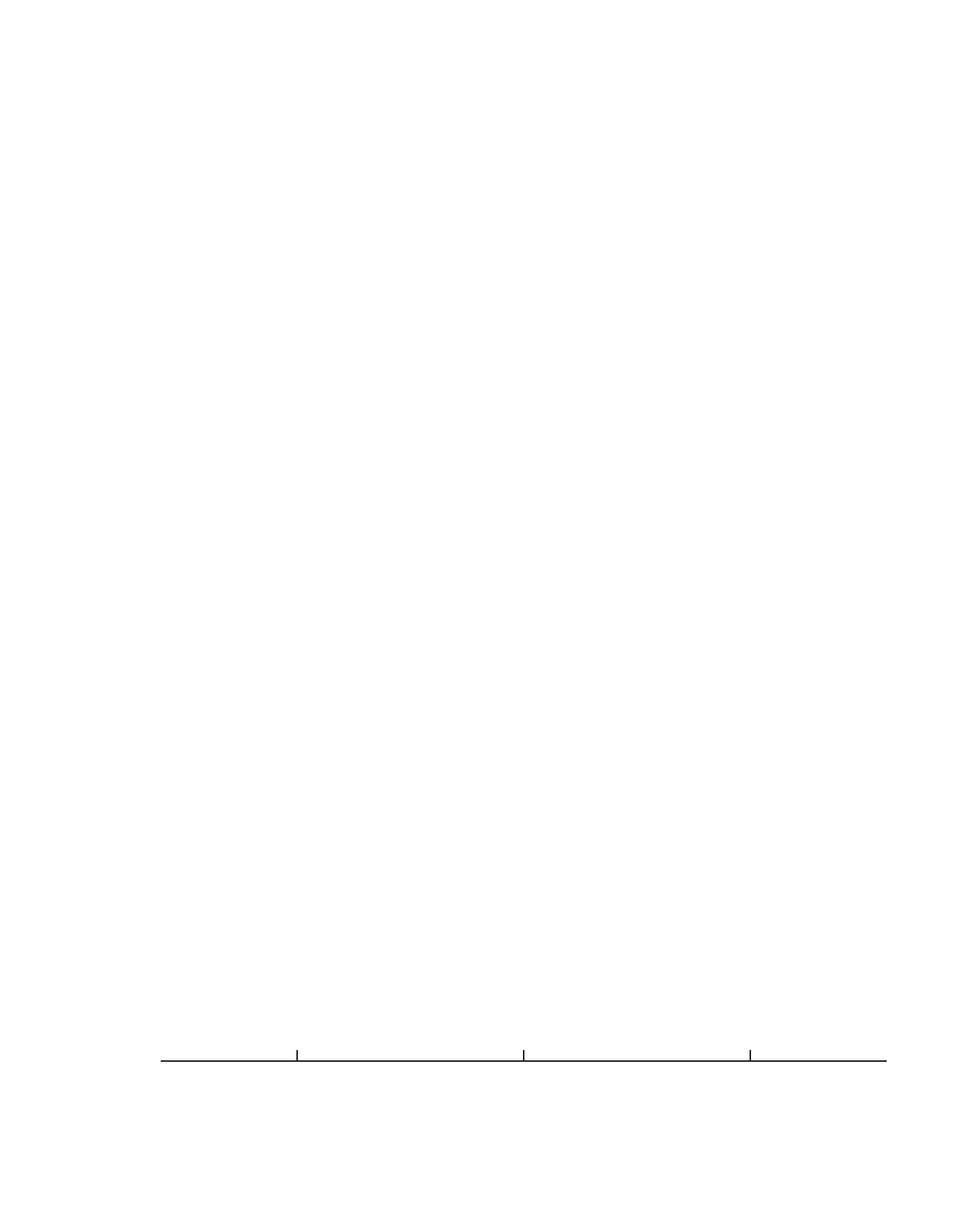}
		\caption{Iterative endpoint fit.}
		\label{fig:ief}
	\end{subfigure}
	\begin{subfigure}{\scanfigwidth}
		\resizebox{\columnwidth}{!}{\Huge 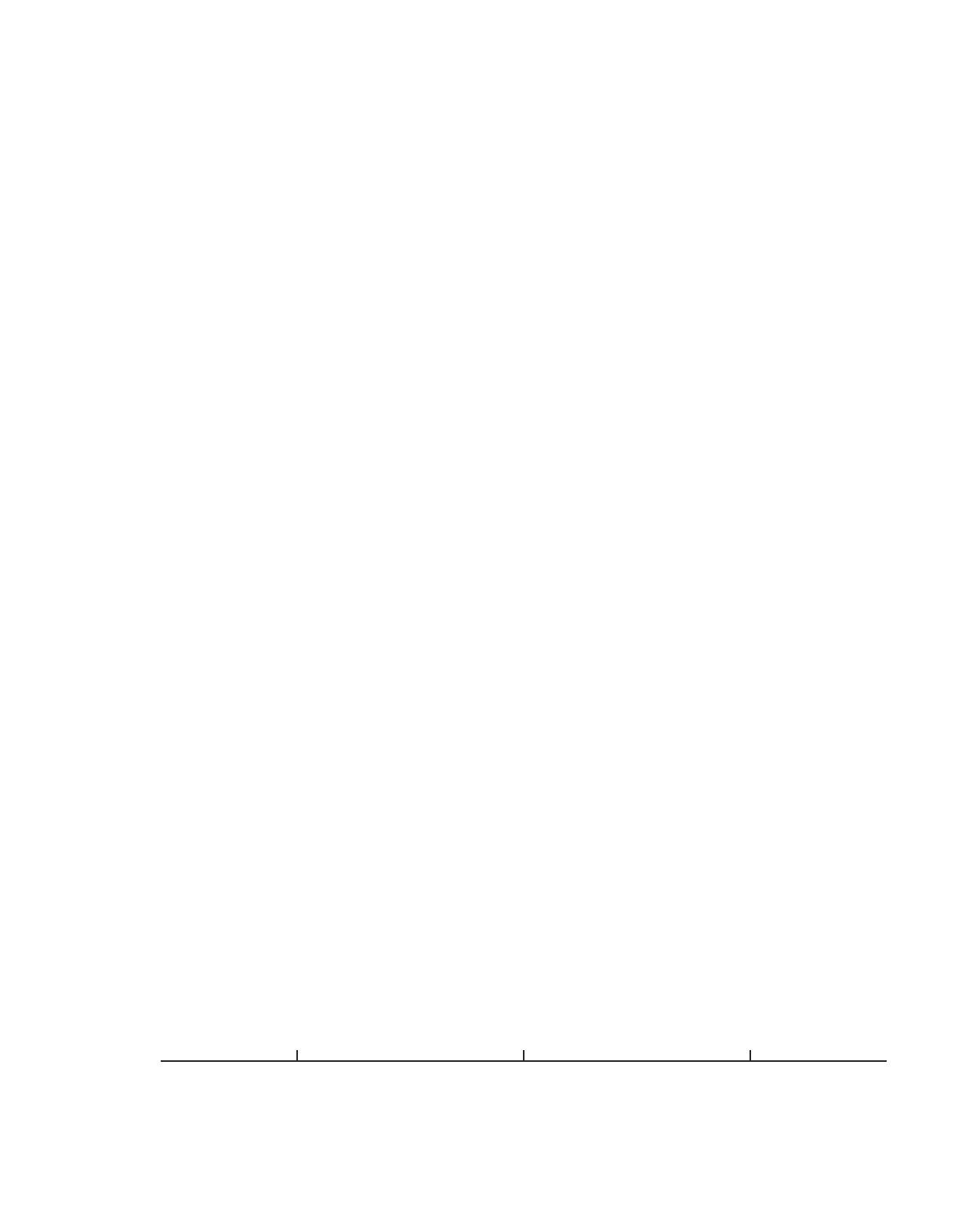}
		\caption{Split-and-merge.}
		\label{fig:sam}
	\end{subfigure}
	\begin{subfigure}{\scanfigwidth}
		\resizebox{\columnwidth}{!}{\Huge 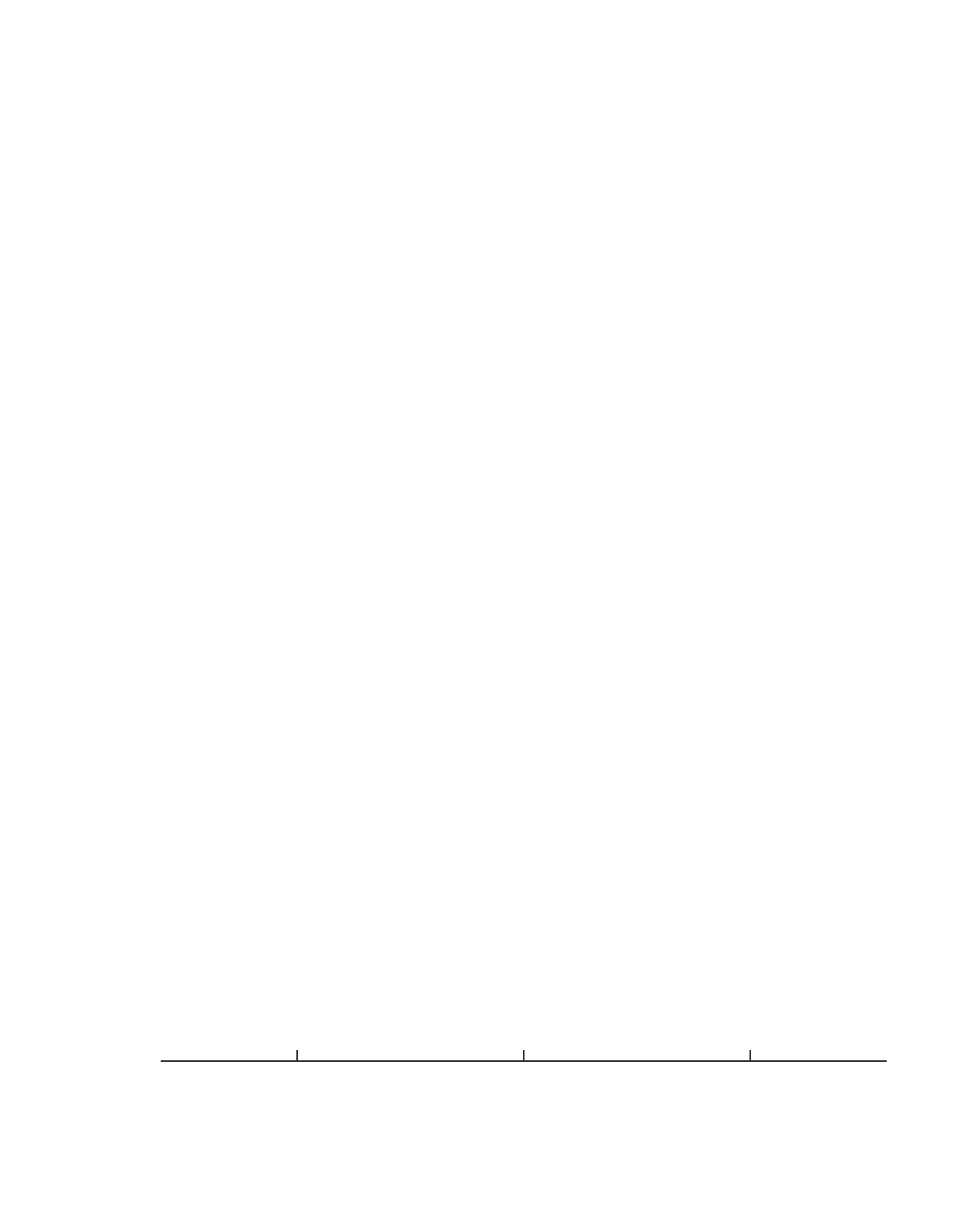}
		\caption{Veeck and Burgard's method.}
		\label{fig:vb}
	\end{subfigure}
	\\
	\begin{subfigure}{\scanfigwidth}
		\resizebox{\columnwidth}{!}{\Huge 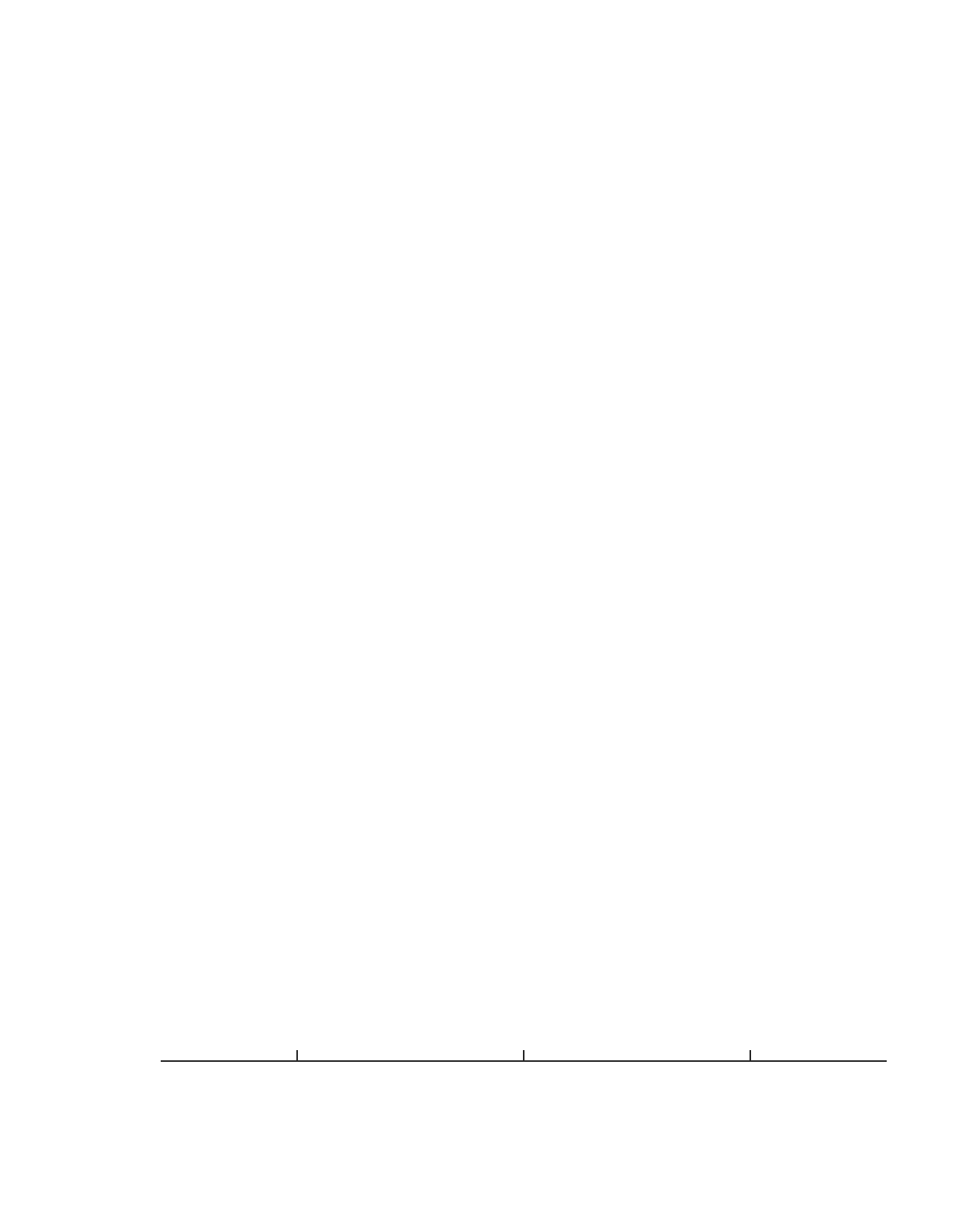}
		\caption{Ours without optimization.}
		\label{fig:ple}
	\end{subfigure}
	\begin{subfigure}{\scanfigwidth}
		\resizebox{\columnwidth}{!}{\Huge 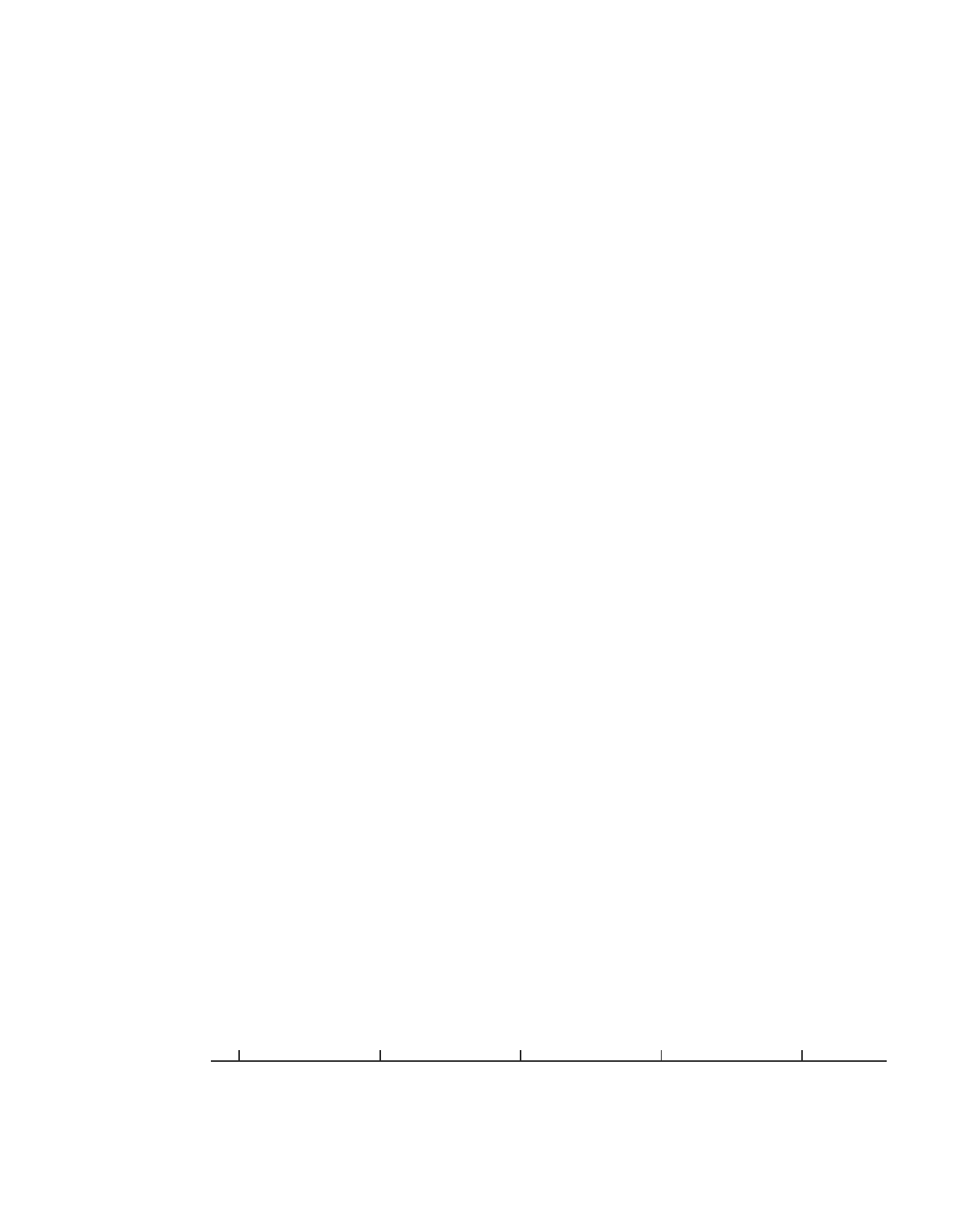}
		\caption{Zoom into (\subref{fig:ple}).}
		\label{fig:plezoom}
	\end{subfigure}
	\begin{subfigure}{\scanfigwidth}
		\resizebox{\columnwidth}{!}{\Huge 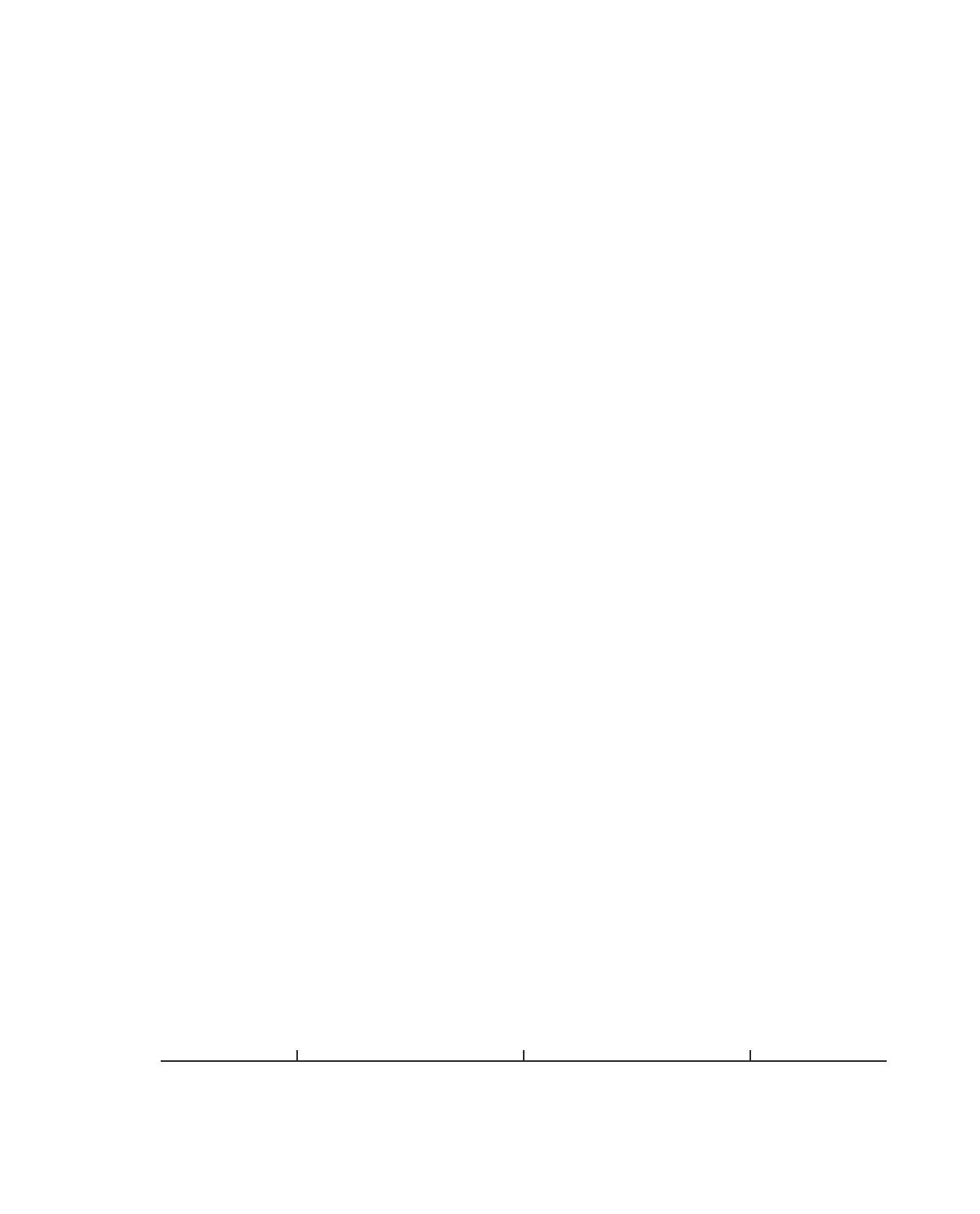}
		\caption{Ours with optimization.}
		\label{fig:plep}
	\end{subfigure}
	\begin{subfigure}{\scanfigwidth}
		\resizebox{\columnwidth}{!}{\Huge 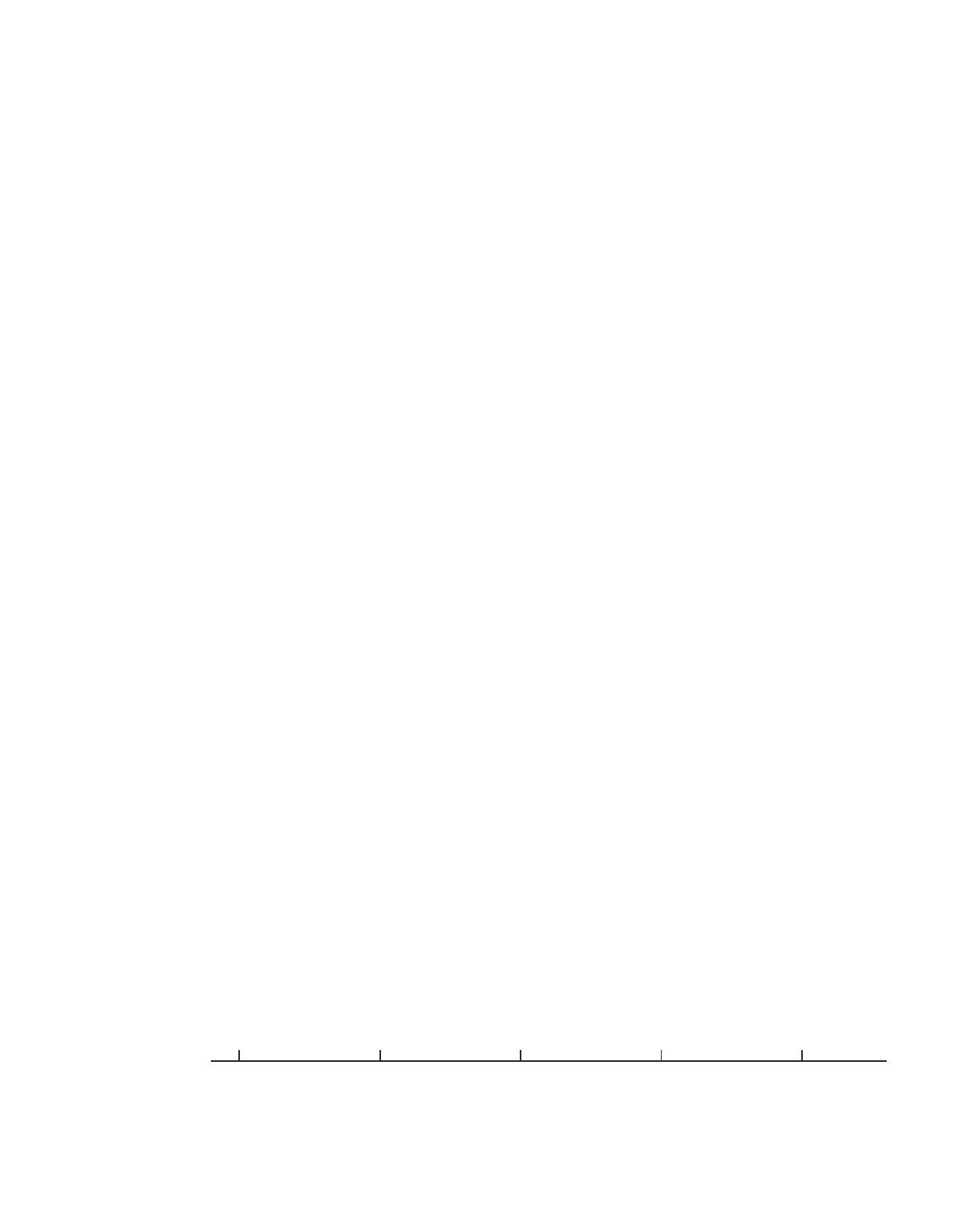}
		\caption{Zoom into (\subref{fig:plep}).}
		\label{fig:plepzoom}
	\end{subfigure}
	\caption{Exemplary results of the various polyline extraction methods applied to the same scan captured in an office.
	All methods respect the requirement of at most \num{10} vertices, except for VB, which produces \num{17} vertices.}
	\label{fig:scan}
\end{figure*}

\begin{figure}
	\begin{subfigure}{\linewidth}
		\centering
		\resizebox{\linewidth}{!}{
			\large
			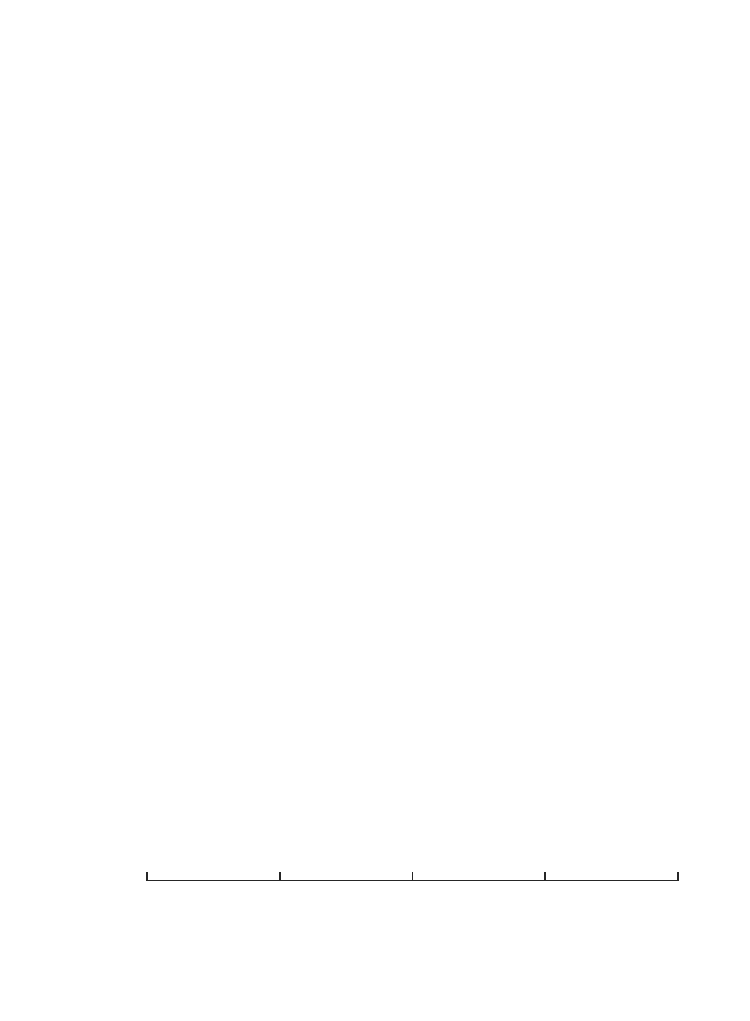
			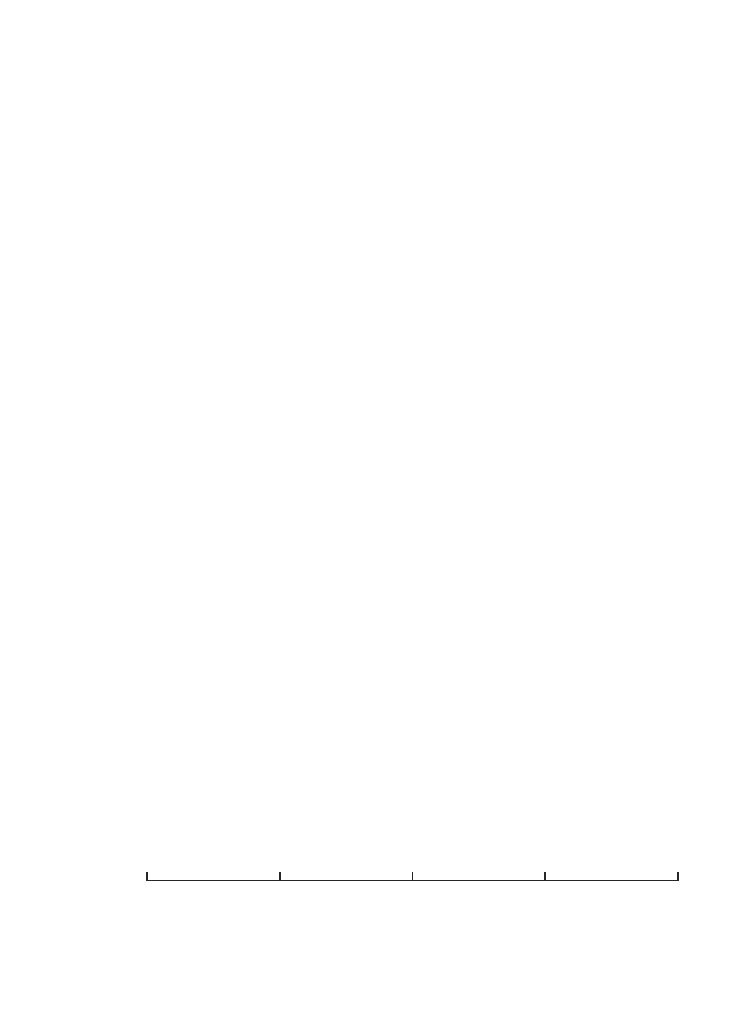
		}
		\caption{Mean accuracy of extracted polylines on real-world lidar data.}
		\label{fig:result_real}
	\end{subfigure}
	\begin{subfigure}{\linewidth}
		\label{fig:rmse}
		\centering
		\resizebox{\linewidth}{!}{
			\large
			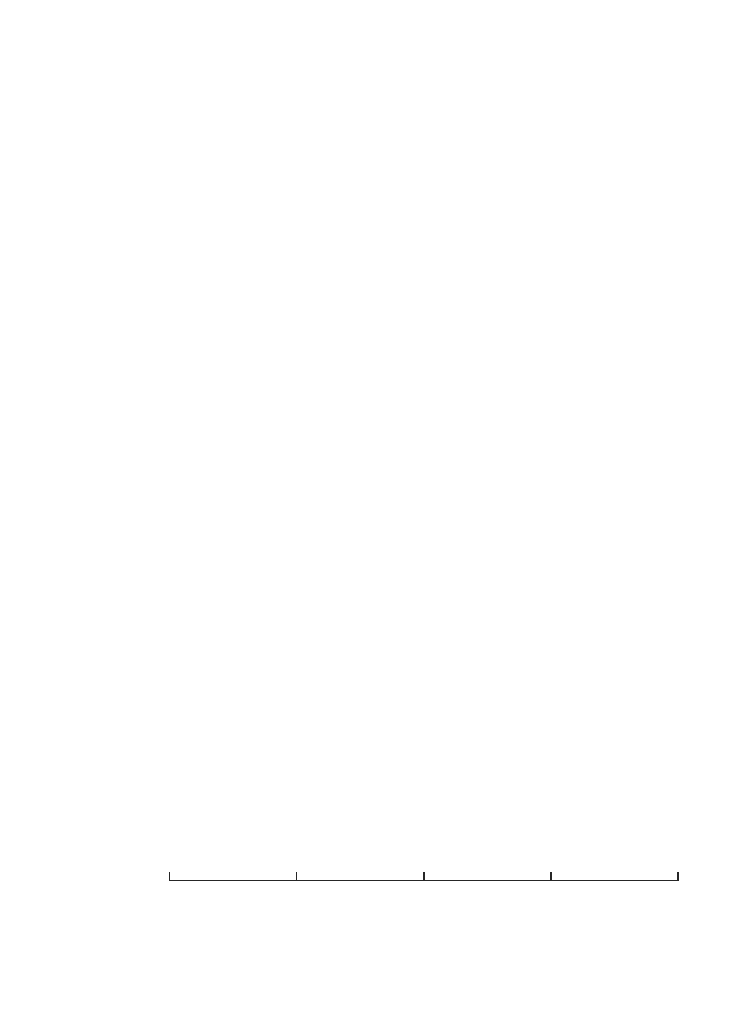
			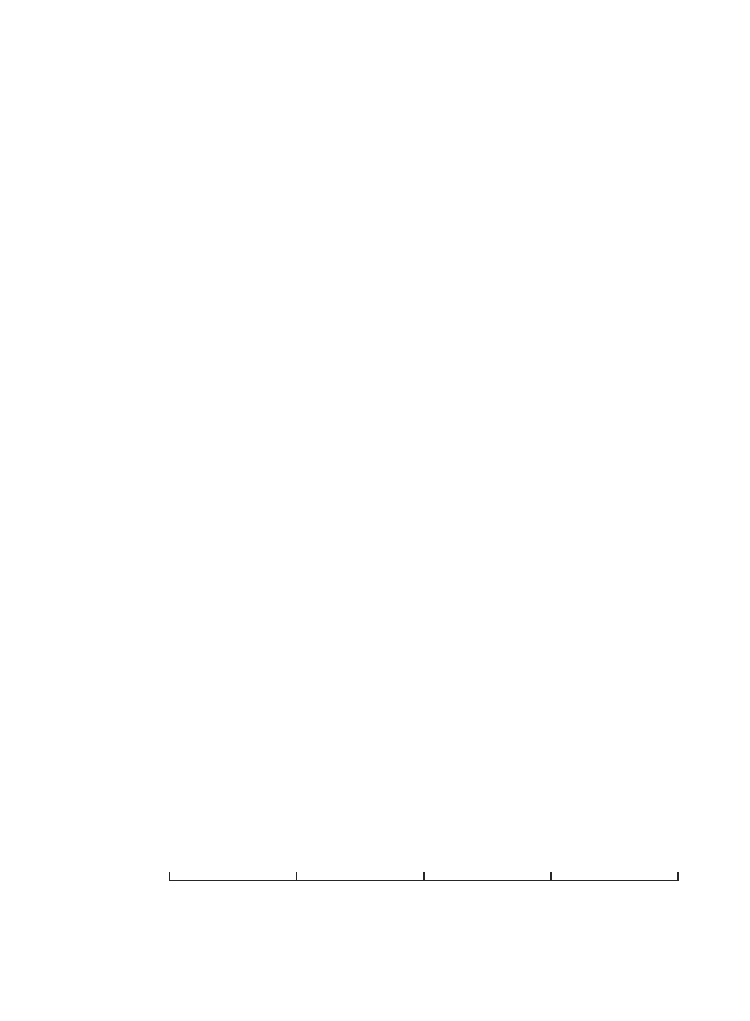
		}
		\caption{Mean accuracy of extracted polylines on simulated lidar data.}
		\label{fig:result_sim}
	\end{subfigure}
	\caption{Experimentally determined accuracies of the investigated polyline extraction methods.
		RMSE denotes the root mean squared error between the measured laser ray endpoints and the hypothetical intersection with the extracted set of polylines, averaged over all scans. 
		The variable~$f$ indicates the fraction of rays explained by the polyline map, whereas $a$ denotes the relative area error between the  polygon extracted from simulated data and the underlying ground-truth polygon.
		The error bars in the plots visualize the standard errors.
		The dashed line on the left side of plot~(\subref{fig:result_sim}) marks the standard deviation of the radial noise for the simulated laser scans.}
	\label{fig:result}
\end{figure}

Figure~\ref{fig:result} summarizes our findings pertaining to the accuracy of the investigated methods.
It displays the evolution of the results over increasing memory footprint, encoded by the number of vertices~$J$.
For each method and each dataset, we evaluate the following values of $J$: \num{10}, \num{20}, \num{30}, \num{40}, \num{50}.
We employ three different metrics to look at the results from different perspectives.
The root mean squared error (RMSE) of the ray radii assesses how closely the extracted polylines represent a scan.
The E in RMSE is the distance between the measured endpoint of a lidar ray and its hypothetical intersection with the polyline, measured along the ray axis.
For each scan, we determine one RMSE value by iterating over all rays.
For every algorithm, we then average the RMSE over all scans to obtain the values shown in figure~\ref{fig:result}.
The second metric, denoted by $f$, quantifies the match between the polyline map and the original scan in a different way.
For each scan, $f$ is the ratio of the number of rays hypothetically intersected by the polylines and the number of rays actually reflected in the measured scan.
The $f$\nobreakdash-values presented in figure~\ref{fig:result} are again averaged over all scans.
For the simulated experimental runs, we introduce the third metric~$a$. 
It measures how well an extracted polyline map recovers the ground truth map.
To determine the $a$\nobreakdash-value, we transform the coordinates of the estimated polyline vertices to polar coordinates with respect to the sensor pose, order them counterclockwise, and build a polygon out of them.
We then compare how well the resulting polygon matches the one that represents the ground-truth.
More specifically, we compute the ratio of areas
\begin{equation*}
a \coloneqq \frac{(a_{\textrm{GT}} \cup a_{\textrm{E}})-(a_{\textrm{GT}} \cap a_{\textrm{E}})}{a_{\textrm{E}}},
\end{equation*}
where $a_{\textrm{GT}}$ denotes the area of the ground-truth polygon, while $a_{\textrm{E}}$ stands for the area of the estimated polygon.

Figure~\ref{fig:result} reveals that although very popular, the split-and-merge algorithm performs poorly relative to the other line extraction methods we selected.
Even more surprisingly, its less elaborate predecessor, iterative endpoint fit, outperforms SAM with respect to every metric we evaluated.
We tracked the reason for this behavior down to the line fitting process:
Especially in the first iterations, when the number of polyline segments is by far too small to represent the structure of the environment, fitting often leads to a degradation of the line estimate.
During the later iterations, when refining the polylines, the algorithm is not able to compensate this inaccurate prior.
The described behavior is not only apparent in the RMSE, but also in the $f$\nobreakdash-value.
While IEF explains all measurements, SAM does not account for a significant amount of rays.
So in contrast to the conclusion that Nguyen et~al.~\cite{nguyen2005} drew after experimentally comparing SAM and IEF on a dataset of \num{100} scans, we find that split-and-merge performs significantly poorer than iterative endpoint fit.

The results of the method developed by Veeck and Burgard are only given for $J \approx 20$.
This is due to the fact that their approach does not allow to set the memory limits of the resulting polyline directly.
Instead, one has to provide a target value for the Bayesian Information Criterion (BIC), which they use to balance the compromise between memory requirements and accuracy.
Unfortunately, even large variations in the BIC value lead to similar vertex counts. 
For that reason, we are not able to evaluate VB over the whole range of $J$.
The results demonstrate that VB achieves good accuracy, but only for the limited set of laser rays it explains.
At \num{0.71}, the $f$\nobreakdash-value turns out comparatively low both on real data and in simulation.
The reason is the grid map-based initialization of the polylines, which discards grid cells with low occupancy values.
Figure~\ref{fig:vb} illustrates this behavior.
In contrast to all other methods, VB is not able to extract the line that runs approximately through coordinate $(10,13)$.
The occupancy probability along this line is simply too small to qualify as an initial polyline.
As a consequence of this behavior, VB returns the most inaccurate polygons in simulation.

Although to our knowledge never evaluated in a robotics context, Visvalingam's algorithm returns comparatively low RMSE values in both experimental settings.
The characteristic it suffers from most is its small $f$\nobreakdash-value -- a consequence of the fact that the removal of straight lines comes at no cost.
Hence, the algorithm discards any solitary line segment in order to decrease the vertex count.
The exemplary output of the Visvalingam method in figure~\ref{fig:vvl} shows exactly this behavior.
The solitary line segments representing the long walls at the top and on the right side of the image had to make way for the nine vertices in the blue polyline.
Some of them are hardly recognizable because the corresponding kinks in the line are so small.
In simulation, where the scan represents a closed polygon, the described effect does not appear, resulting in an $f$\nobreakdash-value of 1.

The line extractors proposed in this paper, PLE and PLE+, outperform all other methods on real data and in simulation.
As shown in figure~\ref{fig:result}, both algorithms result in significantly smaller RMSE values than the other methods, except for VB, which exhibits a slightly lower RMSE on simulated data.
However, VB is unable to accurately recover the simulated polygons, achieving an $f$\nobreakdash-value of only \num{0.71}, while PLE and PLE+ attain \num{1}.
PLE+ always exhibits smaller RMSE values than PLE, because the optimization minimizes exactly this metric.
The superior $a$-values in figure~\ref{fig:result_sim} demonstrate that minimizing the RMSE also leads to an improved representation of the underlying ground-truth map.
Note that in figure~\ref{fig:result_real}, the $f$-values of PLE and PLE+ are exactly the same, because PLE+ does neither change the topology of the polylines extracted by PLE, nor does the optimization allow boundary rays to interfere with rays that account for measurements outside the polyline.

As expected, the RMSEs of all algorithms in figure~\ref{fig:result_sim} approach the standard deviation of the simulated radial sensor noise for large $J$.
PLE+ even falls below this value, an effect that is due to the algorithm overfitting highly articulated polygons to the noise in the scans.
Correspondingly, the area error increases slightly for large $J$.

Lastly, we report on the computational costs for all methods in table~\ref{tab:times}.
Each algorithm ran in a single thread on an Intel Xeon CPU with \SI{2.50}{GHz}.
The bottom-up line simplification algorithms IEF and SAM exhibit slightly decreasing computation times for increasing $J$, because higher $J$\nobreakdash-values mean less simplification steps.
The repeated fitting steps in SAM turn out to be costly: 
In the worst case, SAM is \num{100}~times slower than IEF.
The reason for the constant timing of Visvalingam's algorithm lies in our implementation:
At first, for every polyline in the map, we compute the incremental errors until the line has vanished.
We then order the errors globally, i.e. over all polyline segments, and remove as many vertices as required to meet the specified vertex count.
Despite this simplified implementation, VVL is at least ten times faster as the popular IEF.
Our algorithms PLE and PLE+ are in the mid-range among the investigated methods.
As a result of the optimization via direct search, we find that the complexity of PLE+ grows approximately quadratically in $J$.
At the same time, the advantage gained by the optimization process decreases for high numbers of vertices, as can be read off figure~\ref{fig:result}.
Therefore, we recommend to use PLE+ to extract only few, but highly accurate vertices.
If memory requirements are less strict and timing becomes an issue, PLE is the right choice.

\begin{table}
	\caption{Mean computation times in seconds.}
	\label{tab:times}
	\centering
		\setlength{\tabcolsep}{0.5em}
		\begin{tabular}{|c|c|cccccc|}
			\hline
			& $J$ & VVL & IEF & SAM & VB & PLE & PLE+ \\
			\hline
			\multirow{2}{*}{Real} & \num{20} 
				& \num{0.056}  &  \num{0.27}  & \num{28} &   \num{0.28}$^*\phantom{^*}$  &  \num{1.4}  &  \num{2.3}
			\\
			& \num{50} 
				& \num{0.050}  &  \num{1.12}  & \num{27} 
				&   ---  &  \num{1.3}  &  \num{6.9} 
			\\
			\hline
			\multirow{2}{*}{Sim.} & \num{20} 
				& \num{0.037}  &  \num{0.38} & \num{103} &   
				\num{0.48}$^{**}$  &  \num{2.3}  & \num{11} 
			\\
			& \num{50} 
				& \num{0.037}  &  \num{1.59} & \num{102} 
				&   ---  &  \num{2.2}  & \num{72} 
			\\
			\hline
			\multicolumn{8}{l}{$^*$ $J=18.5$ \quad $^{**}$ $J=20.7$} 
		\end{tabular}
\end{table}

Both our MATLAB implementation of the presented line extraction approach and the scripts used to conduct and evaluate the experiments are publicly available under \mbox{\url{https://github.com/acschaefer/ple}}.

\section{Conclusion and Future Work}
\label{sec:conclusion}

In order to extract polylines from a 2\nobreakdash-D laser scan, one has to answer two questions:
Which polyline reflects which scan endpoints?
And where are the optimal locations of the polyline vertices?
In the present paper, we answer the first question using a greedy algorithm that minimizes the decrease in measurement probability caused by representing individual scan endpoints by line segments.
The answer to the second question is given by a direct search optimizer that moves the vertices in order to maximize the measurement probability.
Extensive experiments on publicly available datasets and simulated data demonstrate that our approach clearly outperforms all four reference approaches.

Due to the promising results, we will build upon the presented approach in the future and extend the line extractor to three dimensions, resulting in a maximum likelihood approach to extract planes from 3-D laser range scans.

\section*{Acknowledgments}

We thank Michael Veeck for kindly supporting us with the implementation of his line extraction method, and Patrick Beeson, Mike Bosse, Dieter Fox, Giorgio Grisetti, Dirk H{\"a}hnel, Nick Roy, and Cyrill Stachniss for providing the datasets.

\bibliographystyle{IEEEtran}
\bibliography{IEEEabrv,line_extraction}

\end{document}